\documentclass[lettersize,journal]{IEEEtran}
\usepackage{amsmath,amsfonts}
\usepackage{algorithmic}
\usepackage{array}
\usepackage[caption=false,font=normalsize,labelfont=sf,textfont=sf]{subfig}
\usepackage{textcomp}
\usepackage{stfloats}
\usepackage{url}
\usepackage{verbatim}
\usepackage{graphicx}
\def\BibTeX{{\rm B\kern-.05em{\sc i\kern-.025em b}\kern-.08em
    T\kern-.1667em\lower.7ex\hbox{E}\kern-.125emX}}
\usepackage{balance}
\usepackage[numbers,sort&compress]{natbib}
\usepackage{orcidlink}
\hypersetup{hidelinks}

\newcommand{\eg}{\textit{e}.\textit{g}.}

\usepackage[mathscr]{euscript}
\usepackage{multirow}
\usepackage{color}
\usepackage{booktabs}

\begin{document}
%%%%%%%%%%%%%%%%%%%%%%%%%%% title
\title{GMSR: Gradient-Guided Mamba for Spectral Reconstruction from RGB Images}
%%%%%%%%%%%%%%%%%%%%%%%%%%% author
%%%%%%%%%%%%%%%%%%%%%%%%%%% author
\author{Xinying Wang, Zhixiong Huang, Sifan Zhang, Jiawen Zhu,  Paolo Gamba, ~\IEEEmembership{Fellow, IEEE}, Lin Feng

\thanks{
This work was supported by National Natural Science Foundation of China (61972064), LiaoNing Revitalization Talents Program (XLYC1806006) and the Fundamental Research Funds for the Central Universities (DUT19RC(3)012).} 
\thanks{
Xinying Wang, Zhixiong Huang and Sifan Zhang are with the School of Computer Science and Technology, Dalian University of Technology, Dalian, 116024, China. Email:wangxinying@mail.dlut.edu.cn; hzxcyanwind@mail.dlut.edu.cn;201981131@mail.dlut.edu.cn. Jiawen Zhu is with the School of Information and Communication Engineering, Dalian University of Technology, Dalian, 116024, China. E-mail: jiawen@mail.dlut.edu.cn; 
Paolo Gamba is with the Department of Electrical, Computer and BiomedicalEngineering, University of Pavia, 27100 Pavia, Italy. E-mail: paolo.gamba@unipv.it;
Lin Feng is with the School of Information and Communication Engineering, Dalian Minzu University, Dalian, 116600, China, and also with the School of Innovation and Entrepreneurship, Dalian University of Technology, Dalian, 116024, China.E-mail:fenglin@dlut.edu.cn. 
\\Corresponding author: Lin Feng.
}}
% \author{TCSVT Anonymous Authors}

%%%%%%%%%%%%%%%%%%%%%%%%%%%% page header
\markboth{}%
{GMSR: Gradient-Guided Mamba for Spectral Reconstruction from RGB Images}

\maketitle
%%%%%%%%%%%%%%%%%%%%%%%%%%%% abstract
\begin{abstract}
Mainstream approaches to spectral reconstruction (SR) primarily focus on designing Convolution- and Transformer-based architectures.
However, CNN methods often face challenges in handling long-range dependencies, whereas Transformers are constrained by computational efficiency limitations.
Therefore, constructing a efficient SR network while ensuring the quality of reconstructed hyperspectral images (HSIs) has become a major challenge.
Recent breakthroughs in the state-space model (e.g., Mamba) have attracted significant attention from natural language processing to vision tasks due to its near-linear computational efficiency and superior performance, prompting our investigation into its potential for SR problems.
To this end, we propose the Gradient-guided Mamba for Spectral Reconstruction from RGB Images, dubbed GMSR-Net. 
GMSR-Net is a lightweight model characterized by a global receptive field and linear computational complexity.
Its core comprises multiple stacked Gradient Mamba (GM) blocks, each featuring a tri-branch structure.
Building upon the efficient global feature representation from the Mamba, we further innovatively propose spatial gradient attention and spectral gradient attention to guide the reconstruction of spatial and spectral cues.
GMSR-Net demonstrates a significant accuracy-efficiency trade-off, achieving state-of-the-art performance while markedly reducing the number of parameters and computational burdens. Compared to existing approaches, GMSR-Net slashes parameters and FLOPs by substantial margins of 10 times and 20 times, respectively.
Code is available at https://github.com/wxy11-27/GMSR.
\end{abstract}

%%%%%%%%%%%%%%%%%%%%%%%%%%%% introduction
\section{Introduction}
\label{sec:introduction}
\IEEEPARstart
{H}{y}perspectral imaging is a technique utilized for acquiring images by capturing spectral information of an object across  multiple contiguous wavelengths. 
Hyperspectral images (HSIs) can provide spectral information beyond the visible range of the human eye, thus making it possible to analyze and identify the properties, composition, and state of objects in greater detail. The fine spectral information of HSIs can substantially enhance the accuracy of feature classification and identification and is widely used for target identification~\cite{tian2024hyperspectral}, environmental monitoring~\cite{li2022exploring, ang2023adaptornas}, and resource investigation~\cite{su2023probabilistic}. However, hyperspectral imaging devices are often expensive, leading to high data acquisition costs, which limits their widespread use in certain fields. Additionally, due to the inherent physical properties of hyperspectral imaging sensors, HSIs always suffer from complex noise, low signal-to-noise ratio, and low spatial resolution, greatly restricting the accuracy improvement of high-level remote sensing tasks. Snapshot Compressive Imaging (SCI) systems and computational reconstruction algorithms~\cite{cai2021learning, cai2022mask, liu2018rank} are effective means for reconstructing three-dimensional hyperspectral image (HSI) cubes. However, these methods still rely on expensive hardware equipment. As a cost-effective alternative, spectral super-resolution or spectral reconstruction (SR) techniques have attracted attention. SR techniques use color information and known spectral features in RGB images to estimate the spectral response curve of each pixel point to generate HSIs. The current SR methods can roughly be divided into sparse dictionary-learning~\cite{gao2020spectral, fotiadou2018spectral} and deep learning-based methods~\cite{zhao2020hierarchical,chen2023spectral}.

\begin{figure}[!t]
	\begin{center}
		\includegraphics[width=0.48\textwidth]{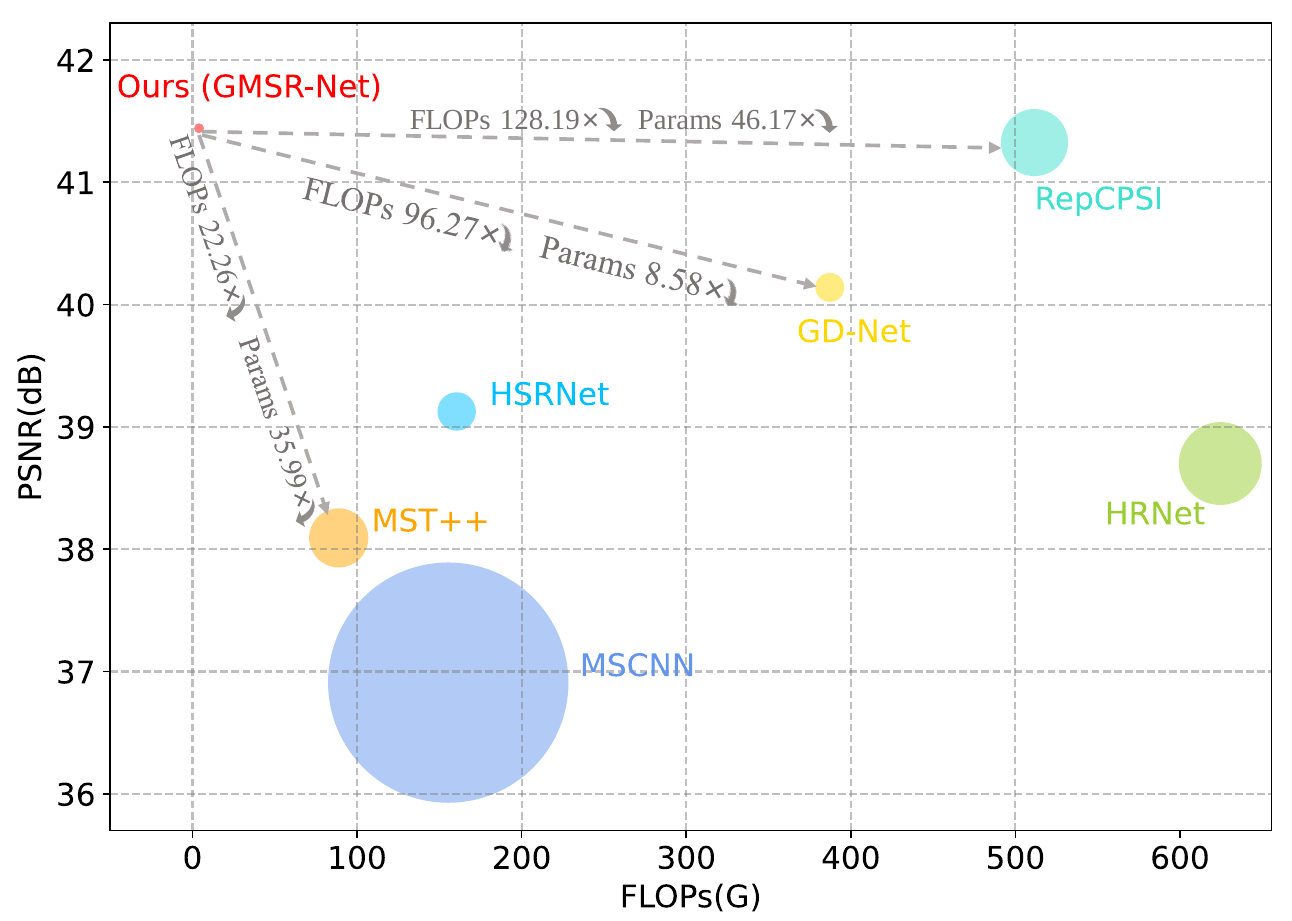}
		\vspace{-1mm}
		\caption{
  %PSNR-Params-FLOPs comparisons with existing spectral reconstruction approaches. The FLOPs (computational complexity) is represented on the horizontal axis, while the vertical axis represents PSNR (performance), with a circle radius indicating Params (memory cost). Our Gradient-Guided Mamba (GMSR-Net) outperforms other methods with lower FLOPs and Params requirements.
  PSNR $v.s.$ Params $v.s.$ FLOPs comparisons with existing spectral reconstruction approaches. For an intuitive analysis, FLOPs and PSNR are represented by the horizontal axis and vertical axis, and the circle radius indicates Params. The proposed Gradient-Guided Mamba (GMSR-Net) outperforms counterparts with dramatically lower FLOPs and Params requirements.
		}
		\label{bubble_chart}
	\end{center}
\end{figure}

The early SR research primarily focused on establishing sparse coding or relatively shallow learning models within specific spectral bands. However, due to their limited expressive power, these methods could only be effective within constrained image domains. In recent years, advancements in deep learning have significantly propelled the development of SR, overcoming many limitations and achieving significant progress. Convolutional neural networks (CNNs) have become the mainstream solution for SR due to their powerful representation and fitting capabilities. Current CNN-based methods treat SR as a regression problem, achieving it by establishing a mapping between HSIs and RGB images. HRNet~\cite{zhao2020hierarchical} uses an encoder-decoder architecture and adds residual global blocks to extend the perceptual range. It won the track 2-Real World in the NTIRE 2020 Challenge Reconstructing Spectra from RGB Images. RepCPSI~\cite{wu2023repcpsi} significantly improves SR accuracy and reduces computational overhead by developing a lightweight coordinate-preserving proximity spectral-aware attention (CPSA) block. While CNN-based SR methods have made impressive progress in performance, CNNs are primarily adept at capturing local features. In the SR task, RGB images provide limited color information, while HSIs contain richer spectral information, which makes it necessary for the model to be able to efficiently map the information in the RGB image into the high-dimensional spectral space. Therefore, the model needs to be able to capture the long-range dependencies between different locations in the input image to ensure spatial consistency and spectral continuity when generating HSIs. In this regard, CNNs appear to be relatively limited.

The Transformer~\cite{vaswani2017attention} model was originally designed for natural language processing (NLP) tasks, but has since been successfully applied to computer vision. Its advantage lies in its ability to handle long-range dependencies and focus on global information. Cai et al.~\cite{cai2022mask} first proposed an end-to-end Transformer-based framework MST, which aims to reconstruct HSI through compressed measurements. 
They proposed spectral-wise multi-head self-attention (S-MSA) to capture inter-spectra similarity and dependencies.
Building on this, Lin et al.~\cite{cai2022coarse} embedded the sparsity of HSI into the Transformer model and developed a fine learning scheme for spectral synthesis imaging. However, since the Transformer needs to compute the attention mechanism between each image patch, the complexity of computation increases exponentially as the resolution increases. Therefore, Transformer models typically have a large number of parameters, which can lead to challenges in training and deployment in resource-constrained situations. Constrained by computational resources, it is usually necessary to divide the image into smaller patches. This pre-processing step inevitably results in each patch contains only a portion of the object, thus providing limited contextual information, which damages the accuracy of reconstruction.

Recently, a novel architecture named Mamba~\cite{gu2023mamba} has attracted widespread attention. Compared to traditional Transformers, 
Mamba demonstrates greater efficiency in handling long sequences while maintaining training and inference effectiveness. 
Scholars have begun to introduce Mamba into the vision fields, including remote sensing image processing, 
achieving innovative progress in tasks such as image classification~\cite{zhao2024rs} and semantic segmentation~\cite{zhu2024samba}. 
However, the aforementioned tasks emphasize high-level information in images, such as objects, scenes, and semantic labels. 
%The capability of Mamba to reconstruct low-level information such as colors, textures, and edges is still underexplored and calls for further investigation. This also inspires us to delve deeper into the potential of Mamba in remote modeling within HSI SR networks.
The potential of Mamba to reconstruct low-level information such as colors, textures, and edges remains largely unexplored and warrants further investigation. This observation also motivates us to explore the capabilities of Mamba in modeling within HSI SR networks.

Mamba requires scanning the image row-by-row or column-by-column, which leads to the separation of tokens that are otherwise adjacent in 2D space, thus increasing the difficulty of feature extraction in localized regions. However, in HSI, local details such as edges and textures are crucial for subsequent analysis and understanding. The gradient map is a great means to effectively reflect the edge and texture information in an image. Currently, the extraction and utilization of image gradients has been widely recognized as a key technique in several fields such as target detection~\cite{canny1986computational} and image segmentation~\cite{arbelaez2010contour}. For example, Canny~\cite{canny1986computational} has become a classical edge detection method based on gradient information by utilizing the gradient information of an image at multiple scales. The SPSR~\cite{ma2020structure} employs an additional gradient branching network to generate a high-resolution gradient map as an additional structural prior, thus alleviating the geometric deformation problem. Distinguished from natural images, HSIs contain not only spatial structural information but also rich spectral information. The gradient map of the spectral dimension can reveal patterns and trends between different wavelengths, assisting models in better understanding and capturing variations among spectral features. This helps improve the model's comprehension and analytical capabilities regarding image content. Currently, the utilization of gradient information in the spectral dimension of HSIs remains relatively limited, and there is still room for further improvement in spectral feature extraction.

% To address the aforementioned challenges, we pioneer the integration of the state space model (SSM) into SR tasks and propose GMSR-Net.
Overall, the current SR task faces the following challenges: CNN- and Transformer-based methods struggle to balance accuracy and efficiency, and geometric distortions often occur during the reconstruction process. In order to more efficiently model the long-range dependencies in HSIs with limited computational resources and mitigate the spatial structure deformation, as well as to better understand and capture the variations and correlations between spectrals, we introduce the state-space model in the spectral reconstruction task for the first time and propose GMSR-Net.
GMSR-Net is an efficient SR model with a global receptive field and linear complexity, consisting of multiple Gradient-Guided Mamba (GM) blocks. Each GM block utilizes a three-branch structure. 
%In addition to efficient global feature extraction using the Mamba block, spatial gradient attention and spectral gradient attention are innovatively proposed to fully maintain structural information in both spatial and spectral dimensions. 
In addition to facilitating efficient global feature extraction through the Mamba block, we propose two innovative attention mechanisms: spatial gradient attention and spectral gradient attention. These mechanisms are designed to meticulously preserve structural information across both spatial and spectral dimensions, enabling SSM to be compatible with SR task.
It is worth mentioning that GMSR-Net successfully balances accuracy and computational efficiency. 
As shown in Fig.~\ref{bubble_chart}, GMSR-Net achieves state-of-the-art (SOTA) performance while significantly reducing parameters and computational overhead. 
The main contributions of this paper are summarized as follows:

%\vspace{-1mm}
\begin{itemize}
\item We propose GMSR-Net, the first Mamba framework designed for SR task. GMSR-Net excels in achieving precise SR through the adept capture of long-range information utilizing a global receptive field. Notably, GMSR-Net strikes a favorable accuracy-efficiency trade-off, outperforming existing well-established methodologies. 
Compared with counterparts, GMSR-Net achieves a remarkable reduction in parameters and FLOPs, achieving a reduction by 10 times and 20 times, respectively.
\item 
We introduce a Gradient Mamba block, distinguished by its innovative tri-branch structure. This design facilitates the efficient extraction of global features while seamlessly integrating spatial gradient attention and spectral gradient attention. By harnessing gradient information, our model achieves a comprehensive understanding of spatial and spectral domain variations, thereby facilitating a more meticulous recovery of spatial and spectral details.
\item 
Extensive experiments on three spectral image reconstruction datasets show that the proposed GMSR-Net outperforms previous approaches in terms of efficiency and accuracy, providing a robust and promising lightweight architecture modeling solution for SR tasks.
\end{itemize}

%%%%%%%%%%%%%%%%%%%%%%%%%%%% related
\begin{figure*}[t]
	\centering
	\includegraphics*[width=1\textwidth]{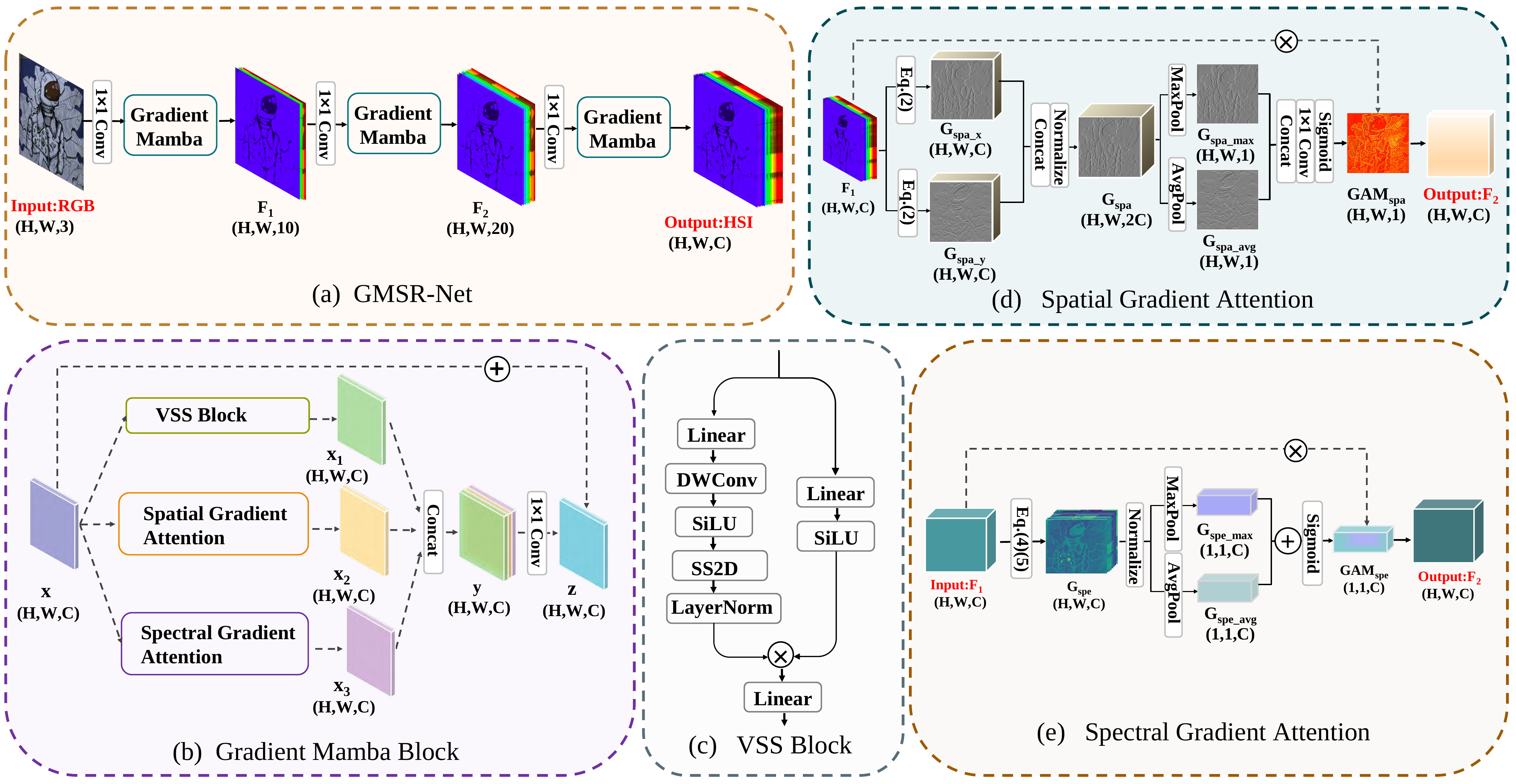}
	%\vspace{-0.25cm}
	\caption{Overview of GMSR-Net. (a) GMSR-Net.(b) Gradient Mamba. (c) VSS Block. (d) Spatial Gradient Attention. (e) Spectral Gradient Attention.}%\vspace{-3.0mm}
	\label{fig:GMSR}
	%\vspace{-2mm}
\end{figure*}

\section{Related Work}
\subsection{Spectral Reconstruction from RGB Images.}
Spectral reconstruction (SR) is an important technique aimed at reconstructing high spectral resolution HSIs only from RGB images. It effectively addresses common challenges in hyperspectral imaging, such as high acquisition cost and low spatial resolution. In recent years, SR techniques have been rapidly developed from linear interpolation to sparse recovery. We summarize the SR algorithms into two categories: a priori-based methods and deep learning-based methods.

Most prior-based methods first perform statistical analysis on the data to obtain the intrinsic properties and features of the image, which are then utilized for SR. Since HSI exists in the form of a data cube, prior knowledge of it typically includes spatial structural similarity, sparsity, and correlations between spectra. Arad et al~\cite{arad2016sparse} proposed a sparse coding-based method to reconstruct HSIs from RGB images. The sparse representation based on a sparse dictionary accomplishes the task of reconstructing individual pixels without considering the spatial correlation, which limits the quality of the reconstructed image. To solve the above problem, Geng et al~\cite{geng2019spatial} proposed a spatially constrained dictionary learning SR method. By spatially introducing contextual information in the sparse representation, the spatial context of pixels is utilized to improve the reconstruction performance. However, HSI requires high spectral information, and the above method's retention of spectral information needs to be enhanced. Given that the spectral signal is usually a relatively smooth function of wavelength, the method~\cite{akhtar2018hyperspectral} employs a Gaussian process to model the spectral signal of each material and combines it with an RGB image in order to obtain a reconstructed HSI. Although this method improves the accuracy of the reconstruction, its solution process is relatively complex.

Deep learning models can be trained in an end-to-end manner, eliminating the need for manually designed feature extractors or post-processing steps. Deep learning-based SR algorithms have exploded in the last five years.
Based on the different reconstruction strategies employed, we can categorize these methods as Unet-like, Residual learning, generative adversarial network, attention, et al~\cite{he2023spectral}. The Unet~\cite{ronneberger2015u} was initially proposed for image segmentation tasks. Due to its fast operation speed and strong feature extraction capabilities, many early SR methods~\cite{stiebel2018reconstructing, yan2020reconstruction} opted to use it as the foundational network. Residual learning can effectively avoid the problem of gradient vanishing or explosion, and many scholars have applied it to the SR task~\cite{gewali2019spectral, kaya2019towards}. Han et al~\cite{han2018residual}. proposed a novel residual CNN-based SR  framework for better recovery of high-frequency content lost by HSI. Since SR can be viewed as an image generation task, utilizing GAN to predict spectral information is a very good way. Zhu et al.~\cite{zhu2021semantic} utilized an adversarial learning approach to achieve excellent reconstruction results without pairs of ground truth HSIs as supervision. The attentional mechanism is a method that mimics the human visual system and is used to increase the model's attention to important parts of the image. Due to the remarkable success of the attention mechanism in tasks such as image segmentation, classification, and target detection, more and more scholars have applied it to SR tasks~\cite{chen2021semisupervised}. Li et al.~\cite{li2021deep} embedded double second-order attention of residual blocks in the network to achieve a more robust representation of spatial and spectral features. 

Although existing deep learning-based methods have achieved satisfactory visualization results, they usually require a large number of parameters and frequent floating-point operations, which increase the computational burden on the processor. In the future, the trend of SR techniques will gradually shift to computational imaging, i.e., in-camera processing. In order to realize in-camera processing, the complexity of the algorithms needs to be reduced and the runtime speed needs to be increased. Therefore, designing a more efficient and concise reconstruction network is a topic worthy of in-depth research.

\subsection{State Space Models}
The state space model (SSM) is a commonly used model in control theory and has recently been innovatively integrated into deep learning. It is utilizing an intermediate state variable such that all other variables are linearly correlated with the state variable and the inputs, thus greatly simplifying the problem. Since typically our inputs are discrete, the Linear State-Space Layer (LSSL)~\cite{gu2021combining} discretizes the continuous-time SSM to obtain two discretized representations. The S4~\cite{gu2021efficiently} proposes to normalize the parameters into a diagonal structure, thus addressing the key obstacles related to the computational and memory efficiency of LSSL. Mamba~\cite{gu2023mamba} is an epic upgrade to S4. It possesses a simple selection mechanism that allows for an RNN mode of performing very fast training. However, the above-mentioned models mainly focus on unidirectional sequential data, primarily dealing with tasks such as language understanding~\cite{ma2022mega} and content-based reasoning~\cite{gu2023mamba}, and cannot handle image data lacking specific directionality. To address the issue of directional sensitivity in the model, Vim~\cite{zhu2024vision} and Vmamba~\cite{liu2024vmamba} transform arbitrary visual images into ordered sequences and introduce a scanning module that traverses the image space domain. While maintaining linear complexity, they preserve a global receptive field. Subsequently, Mamba has inspired numerous pioneering works in fields such as image restoration~\cite{guo2024mambair}, medical image segmentation~\cite{ruan2024vm}, and video understanding~\cite{li2024videomamba}.

Thanks to Mamba's remarkable success in natural and medical image fields, many researchers are now using it for remote sensing image processing. Zhao et al.~\cite{zhao2024rs} proposed RS-Mamba for remote sensing image segmentation. This method employs an omnidirectional selective scan module, allowing selective scanning of remote sensing images in multiple directions. Additionally, Zhu et al.~\cite{zhu2024samba} introduced Samba, a semantic segmentation framework for high-resolution remote sensing images, providing a performance benchmark for mamba-based methods in remote sensing semantic segmentation tasks. So far, SR work based on Mamba has not been widely conducted. This paper aims to explore the potential of Mamba in the field of SR and comprehensively investigate its possibilities as a new generation SR backbone.

%%%%%%%%%%%%%%%%%%%%%%%%%%%% method
\section{Proposed Method}
\subsection{Overall Architecture} 
\label{sec:3.1}
As shown in Fig.~\ref{fig:GMSR}, (a) introduces the proposed GMSR-Net, which is composed of multiple Gradient Mamba (GM) Blocks cascaded together. GMSR-Net takes a three-channel RGB image as input and aims to reconstruct the corresponding HSI. For the input image $RGB\in \mathbb R^{H\times W\times 3}$, firstly, channel upsampling is performed through a $1\times1$ convolution to obtain the feature map $F_1\in \mathbb R^{H\times W\times \frac{C}{n}}$. Next, $F_1$ is input into the GM block (Fig.~\ref{fig:GMSR} (b)). The GM block consists of three branches, which are the VSS (Visual State Space) block (Fig.~\ref{fig:GMSR} (c)), Spatial Gradient Attention block (Fig.~\ref{fig:GMSR} (d)), and Spectral Gradient Attention block (Fig.~\ref{fig:GMSR} (e)). Wherein, the VSS block models long-range dependency relationships, while Spatial Gradient Attention and Spectral Gradient Attention sharpen structural information from all dimensions. Three feature maps $x_1$, $x_2$, and $x_3$ are obtained through the GM block and concatenated to get $y$. Finally, $y$ is channel downsampled and added with $x$ through skip connections to preserve the details and semantic information in the original input data, resulting in feature $z$. 
%GMSR-Net gradually reconstructs spectral information by repeating the above process and ultimately obtains HSI.
GMSR-Net incrementally reconstructs spectral information through the above iterative  processes, ultimately yielding a high-quality HSI.

\subsection{VSS block}
\label{sec:3.2}
When using the Transformer~\cite{vaswani2017attention} model, the computational complexity of its self-attention mechanism increases quadratically with the length of the context, potentially leading to performance bottlenecks when dealing with long sequences. To address this issue, the ViT~\cite{dosovitskiy2020image} model divides the input image into multiple small patches, aiming to reduce computational complexity and enhance feature extraction efficiency. However, this approach of partitioning the image into small patches may limit the model's ability to model long-range relationships across the entire image. Inspired by the success of the Mamba model in modeling long-range relations with linear complexity, we introduce a visual state-space module(VSS) for the spectral recovery task.

As shown in Fig.~\ref{fig:GMSR}(c), following the reference~\cite{liu2024vmamba}, the input feature $x\in \mathbb R^{H\times W\times C}$ is processed through two parallel branches. In one branch, the channels are first expanded to $\lambda C$ via a linear layer, where $\lambda$ is a predefined channel expansion factor. Then, features are extracted using a $3\times3$ convolutional layer and activated through the SiLU activation function. Finally, processing is carried out through a SS2D and layer normalization. The other branch consists of channel expansion, a linear layer, and the SiLU activation function. After aggregating the features from both branches, the channel number is projected back to $C$, generating an output $x_1\in \mathbb R^{H\times W\times C}$ with the same shape as the input. The above process can be represented as:
\begin{equation}
\label{eq1}
\begin{array}{l}
x' = LN[SS2D(SiLU((DWConv(Lin(x)))))]\\
x'' = SiLU(Lin(x))\\
{x_1} = Lin(x' \odot x'')
\end{array}
\end{equation}
Where $Lin$ represents the linear layer, and $DWConv$ represents the depth-wise convolution. $SS2D$ is 2D-Selective-Scan module. $LN$ is layernorm and $\odot$ is Hadamard product.

\begin{figure}[!t]
	\begin{center}
		\includegraphics[width=0.475\textwidth]{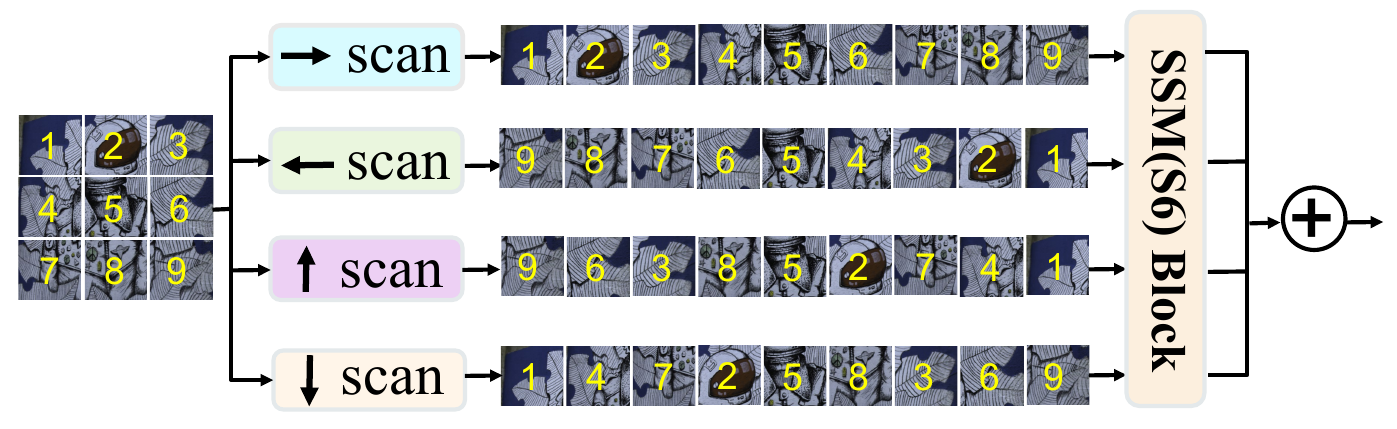}
        \vspace{-3mm}
		\caption{Diagram for the 2D-Selective-Scan (SS2D) Operation.
		}
		\label{fig:SS2D}
	\end{center}
\end{figure}

\textbf{SS2D.} SS2D, the core of the VSS block, draws on the notion of a "selectively scanned state-space sequence model" (S6), which was originally applied to NLP. Unlike traditional attention mechanisms, S6 enables each element in a sequence to interact with any previously scanned sample, thereby reducing the complexity of the attention mechanism from quadratic to linear. 
%However, due to the non-causal nature of visual data, directly applying the S6 method to chunked and flattened visual images can lead to a loss of global receptive fields. 
%To address this issue, we followed the approach outlined in \cite{liu2024vmamba} and adopted a “four-direction” scanning strategy. 
%As depicted in Fig.\ref{fig:SS2D}, to better utilize 2D spatial information, we introduced the 2D-Selective-Scan (SS2D) operation, which flattens the 2D image features into a 1D sequence and scans along four different directions. 
%This allows us to capture the long-range dependencies of each sequence. Finally, all sequences are merged using a summation method and reshaped to restore the 2D structure.
However, the non-causal nature of visual data poses a challenge when applying the S6 method directly to chunked and flattened visual images, potentially resulting in a loss of global receptive fields.
To tackle this issue, we followed the approach outlined in \cite{liu2024vmamba}  and implemented a “four-direction" scanning strategy.
As illustrated in Fig.~\ref{fig:SS2D}, in order to effectively utilize 2D spatial information, we introduced the 2D-Selective-Scan (SS2D) operation. 
This operation flattens the 2D image 
%features 
into a 1D sequences and scans along four different directions, enabling the capture of long-range dependencies within each sequence.
Subsequently, all sequences are merged by summation and reshaped to the original 2D structure.

\subsection{Spatial Gradient Attention}
\label{sec:3.3}
Although flattening spatial data into 1D tokens endowing capturing long-range dependencies, it disrupts the local 2D dependencies of the image, thereby weakening the model's ability to accurately interpret spatial relationships. 
While SS2D partially addresses this issue by scanning images horizontally and vertically, it still struggles to maintain the proximity of original adjacent tokens during the scanning sequence, resulting in the destruction of local details such as edges and textures. 
%To solve the above problems, we propose spatial gradient attention to further deepen the contours in HSI and increase the clarity of HSI.
To solve the above problems, we introduce spatial gradient attention to further enhance the local cues (\eg, contours) in HSIs and improve their clarity.

The structure of Spatial Gradient Attention is illustrated in Fig.~\ref{fig:GMSR}(d).
Define the input feature map as $F_1\in \mathbb R^{H\times W\times C}$. Firstly, employ the first-order finite difference method to obtain the spatial gradient maps of $F_1$ in both horizontal and vertical directions, resulting in ${{\rm{G}}_{{\rm{spa\_x}}}}$ and ${{\rm{G}}_{{\rm{spa\_y}}}}$. Then, concatenate them and perform normalization to obtain ${{\rm{G}}_{{\rm{spa}}}}$. The above process can be represented as:
\begin{equation}
\label{eq2}
\begin{array}{l}
{{\rm{G}}_{{\rm{spa\_x}}}}{\rm{ = }}{{\rm{F}}_{\rm{1}}}\left[ {:,{\rm{ 1}}:,{\rm{ }}:} \right]{\rm{ }} - {\rm{ }}{{\rm{F}}_{\rm{1}}}\left[ {:,{\rm{ }}: - 1,{\rm{ }}:} \right]\\
{{\rm{G}}_{{\rm{spa\_y}}}}{\rm{ = }}{{\rm{F}}_{\rm{1}}}\left[ {1:,{\rm{ }}:,{\rm{ }}:} \right]{\rm{ }} - {\rm{ }}{{\rm{F}}_{\rm{1}}}\left[ {: - 1,{\rm{ }}:,{\rm{ }}:} \right]\\
{{\rm{G}}_{{\rm{spa}}}} = Normalize([{{\rm{G}}_{{\rm{spa\_x}}}},{{\rm{G}}_{{\rm{spa\_y}}}}{\rm{])}}
\end{array}
\end{equation}
where ${{\rm{G}}_{{\rm{spa\_x}}}}$ and ${{\rm{G}}_{{\rm{spa\_y}}}}$ represent the spatial gradient map in the horizontal and vertical directions, respectively. $[.]$ denotes concatenation operation.

Next, we perform channel-wise global max pooling and global average pooling operations on ${{\rm{G}}_{{\rm{spa}}}}$, obtaining two results, ${{\rm{G}}_{{\rm{spa}}}}_{{\rm{\_max}}}$ and ${{\rm{G}}_{{\rm{spa}}}}_{{\rm{\_avg}}}$, respectively. Then, these two results are concatenated along the channel dimension. Subsequently, a convolution operation is applied to reduce the number of channels to 1. Finally, spatial gradient attention feature ${\rm{GM}}{{\rm{A}}_{{\rm{spa}}}}$ is generated through the sigmoid function. Ultimately, ${\rm{GM}}{{\rm{A}}_{{\rm{spa}}}}$ is element-wise multiplied with the input features of the module, yielding spatially enhanced output feature $F_2$.
\begin{equation}
\label{eq2}
\begin{array}{l}
{G_{spa{\rm{\_}}Max}} = MaxPool({G_{{\rm{spa}}}})\\
{G_{spa{\rm{\_}}}}_{Avg} = AvgPool({G_{{\rm{spa}}}})\\
{{\rm{F}}_{\rm{2}}} = {{\rm{F}}_1} \odot (\sigma (Conv([{G_{spa{\rm{\_}}Max}},{G_{spa{\rm{\_}}}}_{Avg}])))
\end{array}
\end{equation}
where $\sigma$ stands for sigmoid operation and $Conv$ represents the convolution operation.

\subsection{Spectral Gradient Attention}
Unlike natural images, HSIs not only contain spatial structural information but also cover tens or even hundreds of spectral channels. By combining these spectral channels, HSIs can provide more detailed and precise color and spectral information, thereby enabling finer object recognition and analysis. Therefore, for HSIs, spectral information is equally important as spatial information. The spectral dimension gradient in HSIs reflects the variation trends of each pixel across different spectral bands. During the process of reconstructing hyperspectral, utilizing the spectral dimension gradient of HSIs can help to extract richer spectral information, and the missing spectral information can be better inferred. Based on this concept, we innovatively propose the spectral gradient attention, which is designed to help the model understand the trend of changes between different channels for more accurate and effective spectral reconstruction.

Fig.~\ref{fig:GMSR}(e) illustrates the spectral gradient attention design. Define $F_1\in \mathbb R^{H\times W\times C}$ as the input feature map. Firstly, compute the gradient map along the spectral dimension using the first-order finite difference method, where the calculation is formulated as:
\begin{equation}
{{\rm{G}}_{{\rm{spe\_paritial}}}}{\rm{ = }}Normalize({{\rm{F}}_{\rm{1}}}\left[ {:,{\rm{ }}:,{\rm{ 1}}:} \right]{\rm{ }} - {\rm{ }}{{\rm{F}}_{\rm{1}}}\left[ {:,{\rm{ }}:,{\rm{ }}: - 1} \right])
\end{equation}

Due to the reduction of one channel after computing the gradient image, to ensure that the number of channels in the ${{\rm{G}}_{{\rm{spe}}}}$ matches that of the input image, we can adopt the following compensation step:
\begin{equation}
{{\rm{G}}_{{\rm{spe}}}}{\rm{ = }}\left[ {{G_{{\rm{spe\_paritial}}}},{G_{{\rm{spe\_paritial}}}}[:,:,C - 1]{\rm{ }}} \right]{\rm{ }}
\end{equation}
Where ${{\rm{G}}_{{\rm{spe}}}}$ represents the compensated spectral gradient image, ${G_{{\rm{spe\_paritial}}}}[:,:,C - 1]{\rm{ }}$ denotes the last channel of the gradient image, and $[.]$ denotes the concatenation operation.

Next, ${{\rm{G}}_{{\rm{spe}}}}$ is subjected to normalization and spatial-wise global max pooling and global average pooling operations, resulting in ${{\rm{G}}_{{\rm{spe\_max}}}}$ and ${{\rm{G}}_{{\rm{spe\_avg}}}}$. Subsequently, they are summed and activated to obtain the spectral gradient attention ${\rm{GA}}{{\rm{M}}_{{\rm{spe}}}}$. Finally, multiplying the input feature $F_1$ by ${\rm{GA}}{{\rm{M}}_{{\rm{spe}}}}$ yields the spectrally enhanced output $F_2$. The above process can be defined as:
\begin{equation}
\begin{array}{l}
{G_{spe{\rm{\_}}Max}} = MaxPool({G_{{\rm{spe}}}})\\
{G_{spe{\rm{\_}}}}_{Avg} = AvgPool({G_{{\rm{spe}}}})\\
{{\rm{F}}_{\rm{2}}} = {{\rm{F}}_1} \odot (\sigma ({G_{spe{\rm{\_}}Max}} + {G_{spe{\rm{\_}}}}_{Avg}))
\end{array}
\end{equation}

\subsection{Loss Function}
GMSR-Net uses L1 loss for network optimization. Despite our network not adopting complex loss function designs, but rather simply using the $L1$ loss function, it has demonstrated satisfactory convergence during training, fully illustrating GMSR-Net's excellent ability to capture image details and structures. Define the reconstructed HIS as $Z\in \mathbb R^{H\times W\times C}$,the reference HSI as $\hat{Z}\in \mathbb R^{H\times W\times C}$. The loss function $L$ is defined as:
\begin{equation}
L = \frac{1}{{HWC}}\sum\limits_{h = 1}^H {\sum\limits_{w = 1}^W {\sum\limits_{c = 1}^C {\left| {{{\hat Z}_{hwc}} - {Z_{hwc}}} \right|}}}
\end{equation}
Where $H$, $W$ and $C$ are the height, width and number of channels of the image respectively. ${\hat Z_{hwc}}$ and ${Z_{hwc}}$ denote the pixel values of the $h$th row, $w$th column, and $c$th channel in the reconstructed and reference images, respectively.
%%%%%%%%%%%%%%%%%%%%%%%%%%%% experiments

\begin{table*}[htbp]
  \centering
  \renewcommand\arraystretch{1.2}
  \caption{Final reconstruction results on the NITER2020-CLEAN, CAVE, and Harvard datasets. The champion and second place for each metric are highlighted in  \textcolor[rgb]{1,0,0}{\textbf{red}} and  \textcolor[rgb]{0,0,1}{\textbf{blue}}, respectively.}
  \setlength{\tabcolsep}{1.45mm}{
    % \begin{tabular}{lcccccccccccccc}
    \begin{tabular}{lllllllllllllll}
    %p{3.5em}p{0.9em}p{0.9em}p{0.5em}p{0.9em}p{0.5em}p{0.5em}p{0.5em}p{0.9em}p{0.5em}p{0.5em}p{0.5em}p{0.9em}p{0.5em}p{0.5em}
    \toprule[1.0pt]
    \multirow{2}[4]{*}{Method} & \multicolumn{1}{c}{\multirow{2}[4]{*}{Params(M)}} & \multicolumn{1}{c}{\multirow{2}[4]{*}{FLOPs(G)}} & \multicolumn{4}{c}{NTIRE2020-Clean} & \multicolumn{4}{c}{CAVE} & \multicolumn{4}{c}{Harvard} \\
\cmidrule(lr){4-7} \cmidrule(lr){8-11} \cmidrule(l){12-15}    \multicolumn{1}{c}{} &       &       & RMSE  & \multicolumn{1}{c}{PSNR} & \multicolumn{1}{c}{ASSIM} & \multicolumn{1}{c}{SAM} & \multicolumn{1}{c}{RMSE} & \multicolumn{1}{c}{PSNR} & \multicolumn{1}{c}{ASSIM} & \multicolumn{1}{c}{SAM} & \multicolumn{1}{c}{RMSE} & \multicolumn{1}{c}{PSNR} & \multicolumn{1}{c}{ASSIM} & \multicolumn{1}{c}{SAM} \\
    \midrule
    MSCNN~\cite{yan2018accurate} & 26.6866 & 155.4068 & \multicolumn{1}{r}{0.0177} & 36.9092 & 0.9860 & 2.3036 & 0.0530 & 27.0281 & 0.8617 & 9.9603 & 0.0017 & 29.9853 & 0.9381 & 10.3051 \\
    HRNet~\cite{zhao2020hierarchical} & 3.1705 & 624.5023 & \multicolumn{1}{r}{\textcolor[rgb]{0,0,1}{\textbf{0.0152}}} & 38.7003 & \textcolor[rgb]{1,0,0}{\textbf{0.9924}} & 1.7965 & 0.0541 & 28.1879 & 0.9425 & 7.8588 & 0.0016 & 30.7942 & 0.9594 & 7.6286\\
    HSRnet~\cite{he2021spectral} & 0.6826 & 160.5390 & \multicolumn{1}{r}{0.1619} & 39.1257 & 0.9869 & 2.0972 & 0.0568 & 27.0178 & 0.8910 & 9.5418 & \textcolor[rgb]{1,0,0}{\textbf{0.0015}} & \textcolor[rgb]{0,0,1}{\textbf{31.0960}} & \textcolor[rgb]{1,0,0}{\textbf{0.9633}} & \textcolor[rgb]{1,0,0}{\textbf{7.4106}} \\
    AGD-Net~\cite{zhu2021deep} & \textcolor[rgb]{0,0,1}{\textbf{0.3861}} & 387.2278 & \multicolumn{1}{r}{0.0154} & 40.1399 & 0.9899 & 1.7361 & 0.0550 & 28.0085 & 0.9191 & 7.6613 & 0.0019 & 27.7370 & 0.8855 & 14.4955\\
    MST++~\cite{cai2022mst++} & 1.6196 & \textcolor[rgb]{0,0,1}{\textbf{88.8684}} & \multicolumn{1}{r}{0.0162} & 38.0922 & 0.9910 & 1.7010 & 0.0740 & 24.6099 & 0.9218 & 13.098 & 0.0018 & 30.0349 & 0.9534 & 9.9057 \\
    RepCPSI~\cite{wu2023repcpsi} & 2.0778 & 511.6173 & \multicolumn{1}{r}{0.0165} & \textcolor[rgb]{0,0,1}{\textbf{41.3243}} & \textcolor[rgb]{0,0,1}{\textbf{0.9919}} & \textcolor[rgb]{0,0,1}{\textbf{1.6551}} & \textcolor[rgb]{0,0,1}{\textbf{0.0469}} & \textcolor[rgb]{0,0,1}{\textbf{29.8896}} & \textcolor[rgb]{0,0,1}{\textbf{0.9566}} & \textcolor[rgb]{0,0,1}{\textbf{6.1905}} & 0.0016 & 30.9306 & 0.9546 &10.0971 \\
    Ours  & \textcolor[rgb]{1,0,0}{\textbf{0.0450}} & \textcolor[rgb]{1,0,0}{\textbf{3.9910}} & \textcolor[rgb]{1,0,0}{\textbf{0.0141}} & \textcolor[rgb]{1,0,0}{\textbf{41.4414}} & 0.9918 & \textcolor[rgb]{1,0,0}{\textbf{1.5425}} & \textcolor[rgb]{1,0,0}{\textbf{0.0464}} & \textcolor[rgb]{1,0,0}{\textbf{30.0539}} & \textcolor[rgb]{1,0,0}{\textbf{0.9598}} & \textcolor[rgb]{1,0,0}{\textbf{5.8133}} & \textcolor[rgb]{1,0,0}{\textbf{0.0015}} & \textcolor[rgb]{1,0,0}{\textbf{31.3022}} & \textcolor[rgb]{0,0,1}{\textbf{0.9613}} & \textcolor[rgb]{0,0,1}{\textbf{7.8511}} \\
    \bottomrule[1.0pt]
    \end{tabular}}
  \label{result}%
\end{table*}%

\begin{figure*}[t]
\begin{center}
\includegraphics[width=1\linewidth]{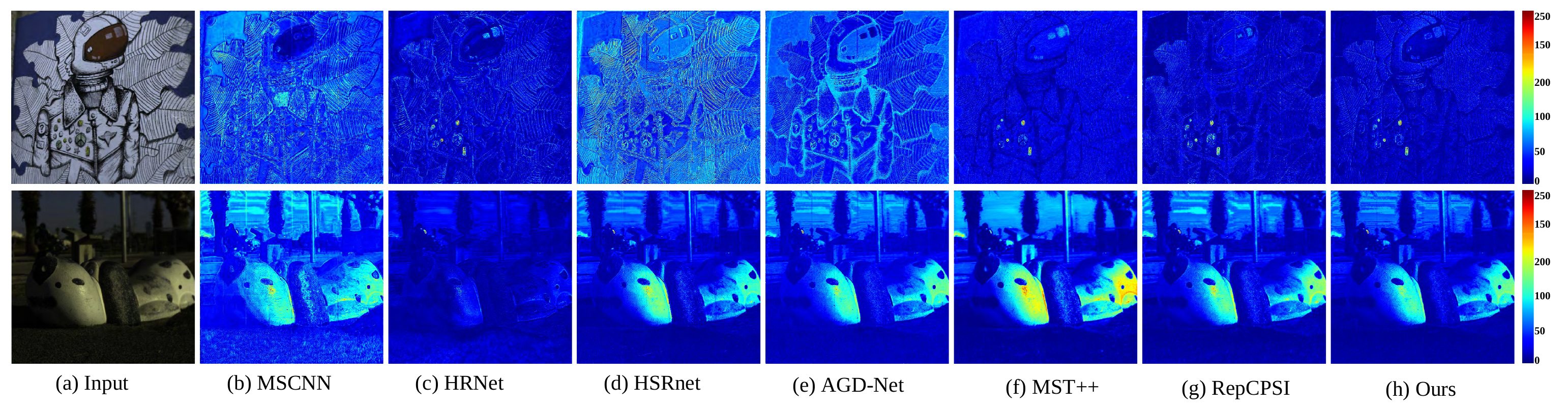}
\vspace{-7mm}
\end{center}
   \caption{Reconstruction results comparison on the NTIRE2020-Clean. The first row is ARAD\_0463 (520 nm) and the second row is ARAD\_0457 (490 nm).}
\label{ARAD}
\end{figure*}

\begin{figure*}[t]
\begin{center}
\includegraphics[width=1\linewidth]{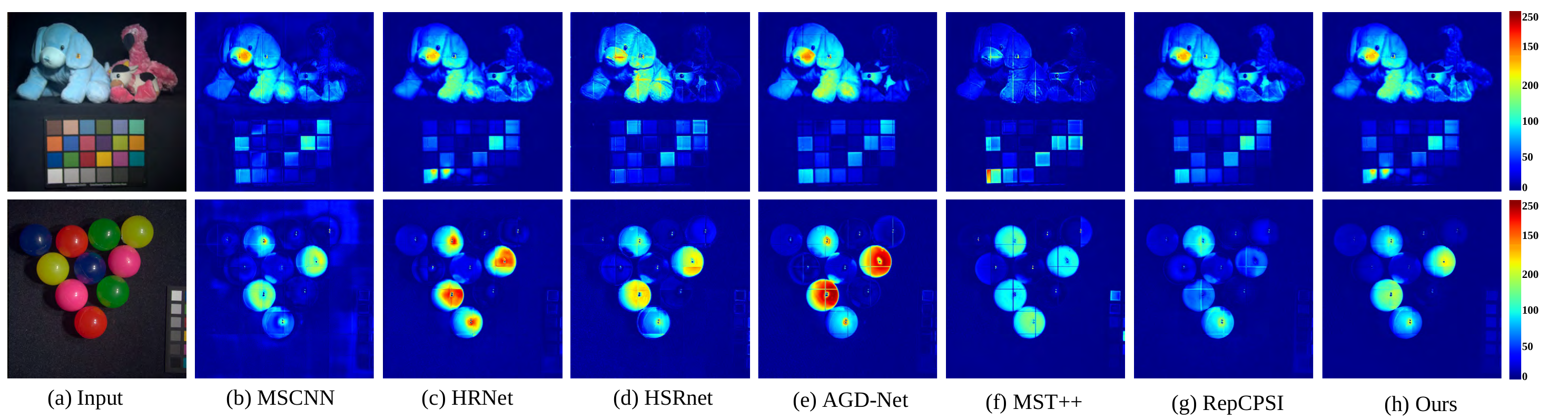}
\vspace{-7mm}
\end{center}
   \caption{Reconstruction results of different methods on the CAVE dataset. The first row is stuffe\_toys (560 nm), the second row is superballs (680 nm).}
\label{cave}
\end{figure*}

\begin{figure*}[t]
\begin{center}
\includegraphics[width=1\linewidth]{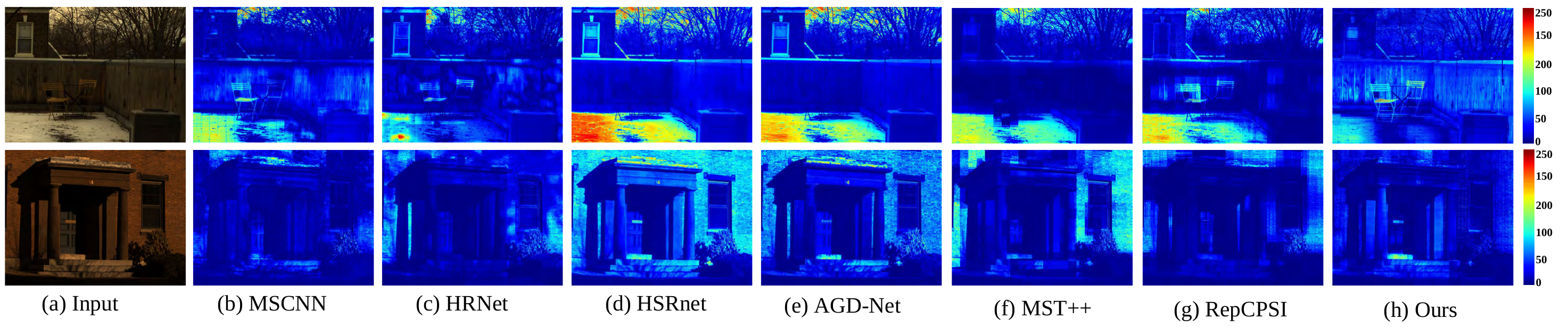}
\vspace{-7mm}
\end{center}
   \caption{Reconstruction results of different methods on the Harvard dataset. The first row is c7 (530 nm), the second row is f7 (680 nm).}
\label{harvard}
\end{figure*}

\begin{figure*}[t]
\begin{center}
\includegraphics[width=1\linewidth]{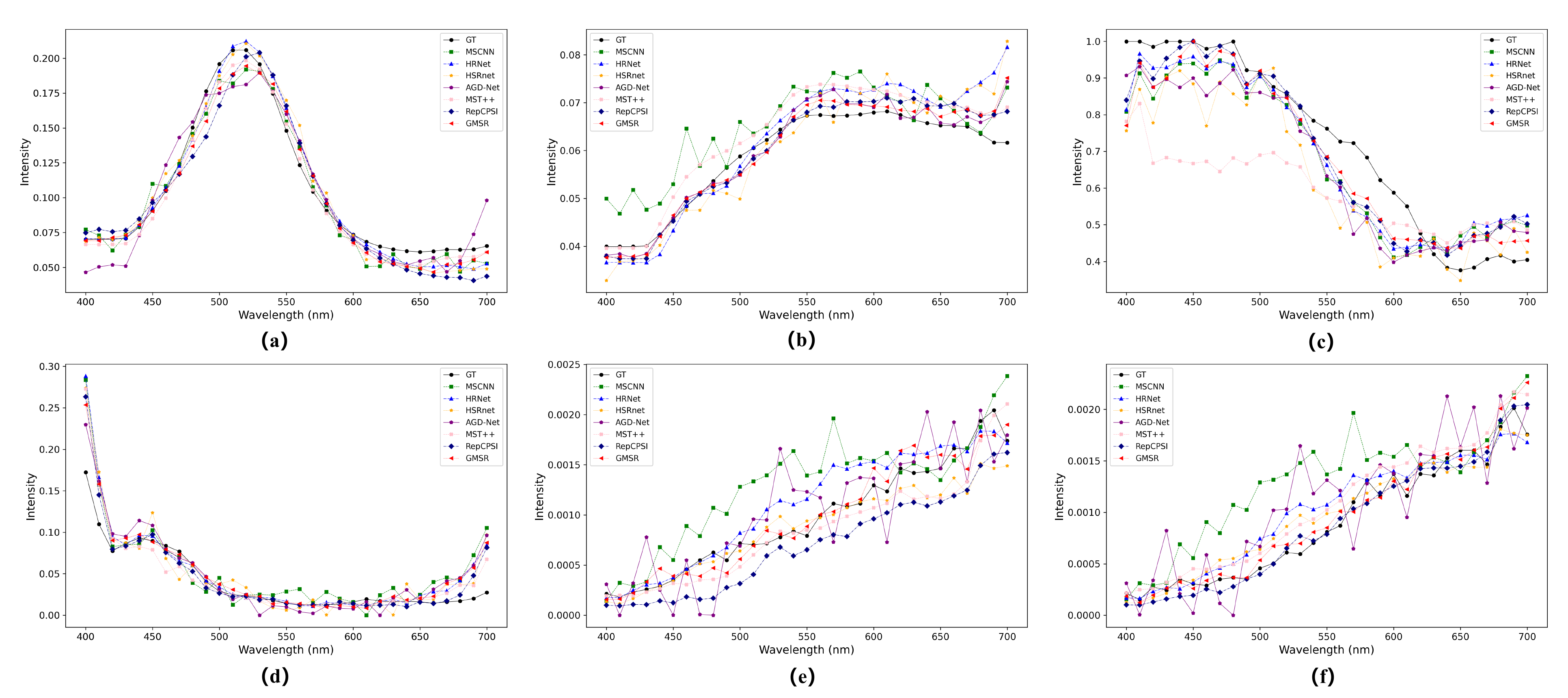}
\vspace{-7mm}
\end{center}
   \caption{ Spectral response curves of randomly selected several spatial points  from the reconstructed HSI of each SOTA method and the groundtruth HSI. (a) and (b) for the NTIRE2020-Clean. (c) and (d) for the CAVE. (e) and (f) for the Harvard.}
\label{spectral_map}
\end{figure*}

\section{Experiments}
\label{Experimentas}
\subsection{Experimental Settiing} 
\subsubsection{Datasets} 
To evaluate the performance of GMSR-Net, we conducted extensive experiments covering three widely used public SR datasets, including NTIRE2020-Clean~\cite{arad2020ntire}, CAVE~\cite{yasuma2010generalized}, and Harvard~\cite{chakrabarti2011statistics}.

\textbf{NTIRE2020-Clean:} The NTIRE2020-Clean dataset is a sub-dataset of the NTIRE2020 Spectral Reconstruction Challenge, designed specifically for evaluating and comparing SR algorithms. The dataset has a spatial resolution of 512 × 482 and covers the wavelength range from 400 to 700 nm, with bands taken at 10 nm intervals,  resulting in a total of 31 bands. The NTIRE2020-Clean dataset consists of 450 training images, 10 validation images, and 20 test images. Since the test set lacks HSIs corresponding to the RGB images, we use the validation set as an independent test set.

\textbf{CAVE:} The CAVE dataset consists of 32 HSIs, with each HSI composed of 31 bands ranging from 400 to 700nm. The corresponding single-band images have a spatial resolution of 512×512. We partition the dataset based on the alphabetical order of image names, with the first 28 images used for training and the remaining 4 images used for testing.

\textbf{Harvard:} The Harvard dataset consists of 50 HSIs, each containing 31 bands. The spectral range covers 420-720nm, with a spatial resolution of 1392×1040. Similar to the CAVE dataset, we partition the dataset based on the alphabetical order of image names, with the first 46 images used for training and the remaining 4 images used for testing.
\subsubsection{ Evaluation Metrics}
To evaluate the performance of the reconstruction model, we used four metrics. These metrics mainly include root mean square error (RMSE), which measure the pixel-level error between the reconstructed HSI and the reference HSI; the peak signal-to-noise ratio (PSNR) and ASSIM, which are used to assess the spatial fidelity; and the SAM, which measures the spectral fidelity. Among them, the smaller the RMSE and SAM, and the larger the PSNR and ASSIM, the higher the quality of the reconstructed HSI.

\subsubsection{Implementation Details}
During the training process, we segmented the original RGB image into patches of size $128 \times 128$. GMSR-Net used the Adam optimizer, where ${\beta _1}=0.9$, ${\beta _1}=0.99$, and $\epsilon=10^{ - 8}$. For better training, we used a learning rate decay strategy, which initialized the learning rate to 0.0001, and set the power = 1.5. The implementation of GMSR-Net is based on the PyTorch framework and was trained on a single NVIDIA 4090Ti GPU for 100 epochs.

\subsection{Comparison With SOTA Methods}
To better demonstrate the effectiveness of the proposed GMSR-Net, we compared it with six methods, namely MSCNN~\cite{yan2018accurate}, HRNet~\cite{zhao2020hierarchical}, HSRnet~\cite{he2021spectral}, AGD-Net~\cite{zhu2021deep}, MST++~\cite{cai2022mst++}, and RepCPSI~\cite{wu2023repcpsi}.

%Tab.~\ref{result} demonstrates the quantitative results of the different methods. It can be seen that the proposed GMSR-Net achieves the best results in most of the metrics, which fully demonstrates the powerful generalization performance of the model. 
Tab.~\ref{result} presents the quantitative results obtained by various methods. It is evident from the table that our proposed GMSR-Net consistently outperforms other methods across most of the evaluation metrics. This outcome underscores the robust generalization capabilities of our model.
On the 2020NTIRE-CLEAN and Cave datasets, the SAM metrics of the GMSR-Net method are reduced by 6.80\% and 6.09\%, respectively, which fully demonstrates the excellent performance of GMSR-Net in spectral reconstruction. Additionally, thanks to the adoption of the spatial gradient attention module, GMSR-Net achieved the best PSNR results on all three datasets, demonstrating excellent spatial reconstruction capabilities.

Fig.~\ref{ARAD} to Fig.~\ref{harvard} provide a spatial comparison of the reconstruction effects achieved by different methods. To better illustrate, we randomly selected different spectral bands from images and displayed the RMSE heatmaps between the reconstructed and ground truth images. The bluer the color, the higher the reconstruction accuracy.
As shown in Fig.~\ref{ARAD}, due to the advantage of spatial gradient attention, GMSR-Net continues to perform well in handling images with complex edge texture structures. Through observation, we found that our GMSR-Net can achieve satisfactory spatial visual effects, which is consistent with the quantitative results.

Fig.~\ref{spectral_map} compares the reconstruction effects of different methods from a spectral perspective. Since continuous spectral curves are commonly used as a means to distinguish the physical and chemical properties of scenes, we randomly selected pixels from the HSI and plotted their spectral response curves. The results indicate that the reconstruction spectra of the proposed GMSR-Net are closer to the ground truth HSI. This is mainly attributed to the accurate portrayal of spectral trends by the spectral gradient attention.

\subsection{Model Efficiency Analyses}
Tab.~\ref{result} shows the FLOPs and parameters of different methods at $512 \times 482$ input size. The results show that the proposed GMSR-Net achieves an excellent trade-off between speed and accuracy in SR task. Compared to the second-best performing method, RepCPSI, GMSR-Net reduces the number of parameters and FLOPs by an astounding $46.17\times$ and $46.17\times128.19$, respectively. In comparison with AGD-Net, which has the second fewest parameters, GMSR-Net reduces the parameter count by $46.17\times8.58$. Similarly, it reduces the FLOPs by $46.17\times5.99$ compared to MST++, the method with the second lowest FLOPS. This improvement is largely due to Mamba’s linear computational complexity when modeling long-range dependencies, allowing for higher efficiency in processing long sequences, especially under limited computational resources. Furthermore, GMSR-Net offers a powerful and promising lightweight backbone for SR, highlighting Mamba’s potential in driving model efficiency and lightweight design.

% Compared with the well-known HRNet, GMSR-Net drastically reduces the number of parameters and FLOPs by a factor of 70 and 40, respectively, while obtaining better reconstruction results. Compared to the latest RepCPSI, this reduction reaches 46 and 128 times, respectively. This demonstrates that GMSR-Net provides a powerful and promising lightweight backbone model for spectral reconstruction tasks, highlighting the potential of Mamba in driving model lightweight.

%%%结构消融实验表
\begin{table*}[]
\centering
\renewcommand\arraystretch{1.2}
\setlength\tabcolsep{6pt}
\caption{Results of Ablation experiments on effects of different ingredients. The best scores are marked in {\textcolor[rgb]{1,0,0}{\textbf{red}}}.}
\begin{tabular}{ccccccccccccc}
\toprule [1 pt]
 & VSS & Transformer & CNN & $G_{spa}$ &$G_{spe}$ &CBAM & RMSE & PSNR & ASSIM & SAM \\ \toprule [1 pt]
GMSR-Net &\checkmark  & && \checkmark&\checkmark & &\textcolor[rgb]{1,0,0}{\textbf{0.0141}}&\textcolor[rgb]{1,0,0}{\textbf{41.4414}}&\textcolor[rgb]{1,0,0}{\textbf{0.9918}}&1.5425\\\hline
Transformer+$G_{spa}$+$G_{spe}$ & &\checkmark& &\checkmark&\checkmark& &0.0150&40.8848&0.9908&1.7452\\ \hline
CNN+$G_{spa}$+$G_{spe}$ &&&\checkmark&\checkmark&\checkmark&&0.0193&38.9331&0.9806&2.0876\\ \hline
VSS+CBAM&\checkmark&&&&&\checkmark&0.0618&39.8417&0.9910&1.6303\\ \hline
VSS+$G_{spa}$&\checkmark&&&\checkmark&&&0.0154&40.4619&0.9911&1.6610\\ \hline
VSS+$G_{spe}$&\checkmark&&&&\checkmark&&0.0156&40.8263&0.9914&\textcolor[rgb]{1,0,0}{\textbf{1.5110}}\\ \hline
$G_{spa}$+$G_{spe}$&&&&\checkmark&\checkmark&&0.0225&37.7243&0.9783&2.1790\\ \hline
VSS&\checkmark&&&&&&0.0197&38.3767&0.9884&1.7210\\ \hline
\toprule [1 pt]
\end{tabular}
\label{Ablation INR}
\end{table*}

\begin{table}[]
\centering
\renewcommand\arraystretch{1.2}
\setlength\tabcolsep{6pt}
\caption{Results of Ablation experiments on the number of Gradient Mamba blocks. The best scores are marked in {\textcolor[rgb]{1,0,0}{\textbf{red}}}.}
\begin{tabular}{ccccccc}
\toprule [1 pt]
n& RMSE & PSNR & ASSIM & SAM & Params(M)  &FLOPs(G)  \\ \toprule [1 pt]
1&0.0188&39.0675&0.9875&2.1168&2.4187&0.0251\\\hline
2&0.0157&40.7241&0.9913&1.5671&3.1603&0.0346\\ \hline
3&\textcolor[rgb]{1,0,0}{\textbf{0.0141}}&\textcolor[rgb]{1,0,0}{\textbf{41.4414}}&\textcolor[rgb]{1,0,0}{\textbf{0.9918}}&\textcolor[rgb]{1,0,0}{\textbf{1.5425}}&3.9910&0.0450\\ \hline
4&0.0157&40.7241&0.9913&1.5671&5.1415&0.0578\\ \hline
5&0.0165&39.5086&0.9906&1.7224&5.7161&0.0657\\ \toprule [1 pt]
\end{tabular}
\label{Ablation n}
%\vspace{-5mm}
\end{table}

\subsection{Ablation Studies}
In order to analyze the role of each component of the GMSR-Net, we performed ablation studies on the NTIRE2020-Clean dataset. The ablation study is divided into two parts, an ablation study for each component of the network and another one for the number of GM blocks.
\subsubsection{Effects of Different Ingredients}
\begin{figure*}[t]
\begin{center}
\includegraphics[width=1\linewidth]{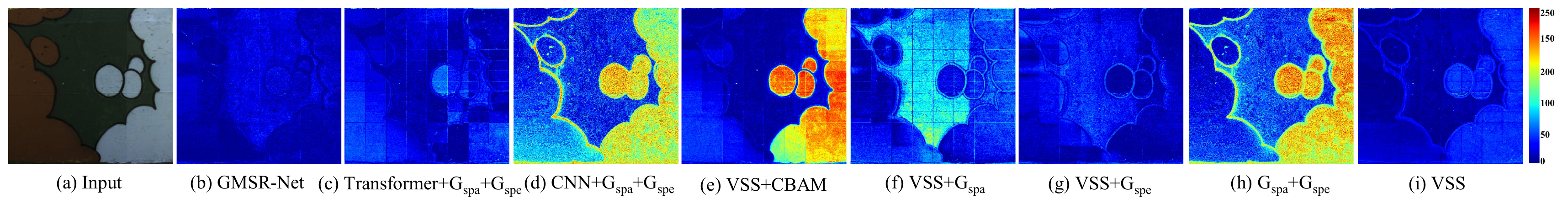}
\vspace{-10mm}
\end{center}
   \caption{Results of ablation experiments on the effects of different ingredients in GMSR-Net. ARAD\_0453 (520 nm) are shown.}
\label{structures}
\end{figure*}

To validate the effectiveness of each component of our proposed GMSR-Net, we conducted the corresponding experiments and show the results in Tab.~\ref{result}. It is clear from the results that the reconstruction results show different degrees of degradation when we use either the Transformer or the CNN instead of VSS block. In particular, when we remove the VSS block in the network, the reconstruction results are significantly degraded. This further demonstrates the excellent characterization ability of Mamba in the reconstruction process, which provides critical support for our model performance. Define $G_{spa}$ and $G_{spe}$ as spatial gradient attention and spectral gradient attention block, respectively. When we replaced $G_{spa}$ and $G_{spe}$ with the spatial and channel attention in CBAM~\cite{woo2018cbam} respectively, we observed a significant decrease in all four metrics. Furthermore, removing $G_{spa}$ and $G_{spe}$ separately resulted in deteriorated reconstruction effects. Importantly, simultaneous removal of $G_{spa}$ and $G_{spe}$ led to a significant decrease in model performance. This result fully illustrates the importance of $G_{spa}$ and $G_{spe}$, which play a key role in maintaining the spatial-spectral details and have a significant impact on improving the reconstruction results. Fig.~\ref{structures} illustrates the visual results of the above experiments. It can be seen that, compared to (b) and (i), the edges are significantly sharper, benefiting from the excellent edge sharpening capabilities of $G_{spa}$ and $G_{spe}$. Additionally, compared to (c) and (d), the overall error is noticeably smaller, %fully 
demonstrating the outstanding global information recovery ability of the VSS block.

\begin{figure*}[t]
\begin{center}
\includegraphics[width=0.8\linewidth]{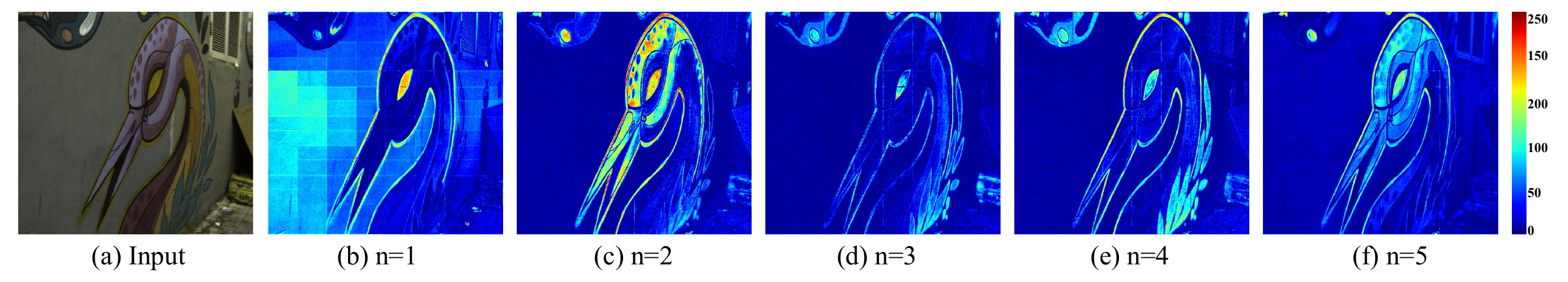}
\vspace{-5mm}
\end{center}
   \caption{Results of ablation experiments on the number of Gradient Mamba blocks. ARAD\_0464 (540 nm) are shown.}
\label{ablation fig n}
\end{figure*}
\subsubsection{Explore the Number of Gradient Mamba blocks}
GMSR-Net consists of multiple GM blocks cascaded together. To investigate the effect of the number of GM blocks on the reconstruction results, we performed an ablation study. Tab.~\ref{Ablation n} and Fig.~\ref{ablation fig n} show the experimental quantitative and visual results. 
%Defining n as the number of GM blocks, the experimental results show that as n increases, the reconstruction results become better and then worse. 
%When the number of GM blocks is 3, the model achieves a good balance between accuracy and efficiency. Fig.\ref{ablation fig n} illustrates the visual results, showing a general alignment with the quantitative findings. Therefore, we finally chose to set the number of GM blocks to 3.
By defining n as the number of GM blocks, our experimental findings reveal a trend wherein the quality of reconstruction initially improves with increasing n, but eventually deteriorates. Notably, when n is set to 3, the model strikes a favorable balance between accuracy and efficiency. This observation is further corroborated by Fig.~\ref{ablation fig n}, which visually depicts results consistent with our quantitative analyses. Consequently, we opt to fix the number of GM blocks at 3 for optimal performance.

%%%%%%%%%%%%%%%%%%%%%%%%%%%% conclusion
\section{Conclusion}
%This paper introduces GMSR-Net, a novel lightweight spectral reconstruction network that leverages the Mamba architecture to model long-range dependencies. 
This paper propose GMSR-Net, a novel lightweight spectral reconstruction model equipped with the long-range dependency-aware vision Mamba architecture.
GMSR-Net subtly incorporates spatial and spectral gradient attention mechanisms to enhance detail recovery during the reconstruction process. 
Our study underscores the potential of the Mamba model in addressing spectral reconstruction challenges and emphasizes the significance of gradient information in this process. 
The proposed GMSR-Net offers a powerful and promising lightweight solution for spectral reconstruction. 
In the future, we will further 
%explore 
%optimizing and 
extend this framework for broader applications in spectral reconstruction and related fields.

\bibliographystyle{IEEEtran}
\bibliography{egbib}

%Biography text here without a photo.

%\begin{IEEEbiography}[{\includegraphics[width=1in,height=1.25in,clip,keepaspectratio]{fig1.png}}]{IEEE Publications Technology Team}
%In this paragraph you can place your educational, professional background and research and other interests.\end{IEEEbiography}

\end{document}